\documentclass[letterpaper]{article} 
\usepackage{openrtag_arxiv}
\nocopyright
\usepackage{times}  
\usepackage{helvet}  
\usepackage{courier}  
\usepackage[hyphens]{url}  
\usepackage{graphicx} 
\usepackage{natbib}  
\usepackage{caption} 
\usepackage{booktabs}
\usepackage{amsmath}
\usepackage{amsfonts}
\usepackage{array}
\usepackage{colortbl}
\usepackage{pifont}

\definecolor{TableHeader}{RGB}{232,241,247}
\definecolor{TableBand}{RGB}{247,250,252}
\definecolor{TableRule}{RGB}{76,132,156}
\definecolor{TableBlue}{RGB}{238,246,255}
\definecolor{TableGreen}{RGB}{239,249,242}
\definecolor{TableOrange}{RGB}{255,244,235}
\definecolor{TablePurple}{RGB}{246,241,255}

\title{OpenRTAG: A Comprehensive Benchmark for Robust Text-Attributed Graph Learning under Data Quality Degradation}
\author{
  Yuze Dai\equalcontrib, Zhihan Zhang\equalcontrib, Yan Zhao, Ruoyu Wu, Xunkai Li,\\
  Zekai Chen, Qiangqiang Dai, Hongchao Qin, Ronghua Li
}
\affiliations{
  Beijing Institute of Technology\\
  \texttt{\{1120240577, 3220241443, 1120243657\}@bit.edu.cn, wrylht@gmail.com}\\
  \texttt{cs.xunkai.li@gmail.com, zackchen02@163.com, qiangd66@gmail.com}\\
  \texttt{qhc.neu@gmail.com, lironghuabit@126.com}
}
\begin{document}

\maketitle

\begin{abstract}
\emph{Text-attributed graphs} (TAGs) are an important graph data form that combine relational structure with rich node text. However, real-world TAGs are often imperfect, with quality issues arising from \emph{text}, \emph{structure}, and \emph{labels}, and typically manifesting as \emph{sparsity}, \emph{noise}, and \emph{imbalance}. These dimensions define nine representative degradation scenarios that can substantially affect TAG learning. Although prior studies have explored specific mitigation strategies, existing evidence remains fragmented across degradation types, datasets, tasks, and model families, leaving TAG robustness insufficiently understood. To address this gap, we present \textbf{OpenRTAG}, a robustness benchmark for text-attributed graph learning. OpenRTAG organizes TAG quality issues into a unified $3\times3$ taxonomy and supports standardized evaluation across nine TAG datasets and three downstream tasks. It systematically evaluates scenario validity and model sensitivity, compares traditional GNNs, LLM-GNNs, and a representative GFM, investigates the effectiveness, efficiency, and robustness of scenario-matched baselines, and further examines model behavior under composite degradation scenarios. OpenRTAG provides a standardized testbed for understanding robustness in TAG learning under realistic low-quality settings.
\end{abstract}

\section{Introduction}

\emph{Text-attributed graphs} (TAGs) have become an important data form for graph learning because they couple relational structure with rich node-associated text~\cite{rethinkingtag,zhang2024tagrl}, supporting applications such as academic networks, social platforms, e-commerce systems, and knowledge-intensive information networks. However, real-world TAGs are often imperfect: their quality issues arise from three coupled modalities, namely \emph{text}, \emph{structure}, and \emph{labels}, and typically manifest as \emph{sparsity}, \emph{noise}, and \emph{imbalance}. Together, these dimensions define nine representative degradation scenarios that can substantially affect TAG learning. Specifically, text sparsity or corruption weakens semantic features, structural sparsity or noise distorts neighborhood context and graph-text alignment, and label sparsity, noise, or imbalance degrades supervision signals. Such challenges affect not only traditional GNNs~\cite{kipf2017semi,velickovic2018gat,hamilton2017inductive} and LLM-GNNs~\cite{tape,zhu2024engine,zhang2025ultratag,ren2024survey}, but also recent \emph{graph foundation models} (GFMs)~\cite{tsgfm,liu2023towardsgfm}, whose robustness under incomplete, noisy, and skewed TAG settings remains insufficiently understood. Motivated by this gap, we study TAG robustness through a benchmark that systematically covers all nine degradation scenarios.


A number of methods have been proposed to mitigate specific data-quality problems, including text denoising and completion~\cite{sun2019ctdmlm,zhang2025ultratag}, graph structure learning~\cite{gslsurvey,llata,li2022stable,liu2022sublime}, label correction, imbalance-aware training~\cite{iglbench,wang2024noisygl,qin2025iglbench}, and LLM-assisted graph modeling~\cite{llm4rgnn,li2024graph}. Yet the current evidence remains fragmented. Existing studies are often tied to a specific degradation type, a small set of datasets, a particular backbone, or a single downstream task; as a result, they do not provide a unified experimental basis for understanding robustness in TAG learning. More importantly, they leave several benchmark-level questions insufficiently answered: \ding{182} Are the constructed degradation scenarios valid, meaningful, non-collapsing, and practical to instantiate? \ding{183} How sensitive are GNNs, LLM-GNNs, and GFMs to different degradation modalities? \ding{184} Which text-, structure-, and label-oriented baseline methods are truly effective under matched low-quality scenarios? \ding{185} And how well do existing methods perform under composite degradation scenarios?


To answer these questions, we present \textbf{OpenRTAG}, a robustness benchmark for text-attributed graph learning. OpenRTAG organizes TAG quality issues into a unified taxonomy over three modalities and three degradation types, yielding nine representative scenarios. It systematically evaluates scenario validity and model sensitivity across multiple TAG datasets and three downstream tasks, including node classification, node clustering, and link prediction, while comparing traditional GNNs, LLM-GNNs, and a representative GFM. It also investigates the effectiveness, efficiency, and robustness of scenario-matched baselines for text, structure, and label degradation, and further examines model behavior under composite degradation scenarios. Rather than serving as another clean-data leaderboard, OpenRTAG provides a standardized testbed for understanding where TAG learning systems fail, which repair strategies help, and how robustness conclusions vary across scenarios, datasets, tasks, and model paradigms.


\textbf{Our Contributions.}
(1) \textit{\underline{Unified degradation taxonomy}}. We define a $3\times3$ benchmark space for TAG quality issues, spanning text, structure, and label modalities under sparsity, noise, and imbalance. 
(2) \textit{\underline{Standardized benchmark framework}}. We build OpenRTAG, a unified framework for scenario construction, data preparation, model and baseline evaluation, and analysis across multiple datasets and three downstream tasks. 
(3) \textit{\underline{Systematic robustness study}}. We evaluate scenario validity and model sensitivity, compare representative GNNs, LLM-GNNs, and GFMs, study the effectiveness, efficiency, and robustness of scenario-matched baselines, and examine model behavior under composite degradation scenarios.

\section{Problem Setting and Degradation Taxonomy}

\subsection{Text-Attributed Graphs}
We consider a \emph{text-attributed graph} (TAG) as a graph in which each node is associated with node-level text~\cite{rethinkingtag}. Formally, a TAG is denoted as $G=(V,E,T,Y)$, where $V$ is the node set, $E$ is the edge set, $T=\{t_i\}_{i\in V}$ denotes node-associated texts, and $Y$ denotes available supervision signals when labels exist. Thus, a TAG contains three coupled information sources: textual semantics $T$, relational structure $(V,E)$, and supervision $Y$.

\subsection{Nine Quality-Degradation Scenarios}
OpenRTAG focuses on data-quality failures from three information sources in TAG learning: \emph{text}, \emph{structure}, and \emph{labels}. Text degradation affects node semantics, structural degradation affects relational context, and label degradation affects supervision. Each modality is organized by three degradation types: \emph{sparsity}, \emph{noise}, and \emph{imbalance}, corresponding to missing information, corrupted information, and uneven quality or support, respectively.
Let the modality and degradation-type sets be:
\begin{equation}
  \begin{aligned}
  \mathcal{M} &= \{\mathrm{text},\mathrm{structure},\mathrm{label}\},\\
  \mathcal{D} &= \{\mathrm{sparsity},\mathrm{noise},\mathrm{imbalance}\}.
  \end{aligned}
\end{equation}
The benchmark scenario space is then defined as:
$\mathcal{S}=\mathcal{M}\times\mathcal{D},$
which yields nine representative degradation scenarios. Given a clean TAG $G$, a scenario $s=(m,d)\in\mathcal{S}$ is instantiated by a scenario constructor $\mathcal{A}_{s,\alpha}$ with perturbation strength $\alpha$:
\begin{equation}
  G_{s,\alpha}=\mathcal{A}_{s,\alpha}(G).
\end{equation}

\textbf{Text degradation.}
Let $\ell_i=|t_i|$ denote the length of node text and let $\phi(t_i)$ denote its semantic content. Text sparsity removes or truncates node text, text noise corrupts tokens or inserts irrelevant content, and text imbalance makes text quality uneven across nodes or classes:
\begin{equation}
  \begin{aligned}
  \mathrm{T\mbox{-}Spa}:&\quad t_i'=\emptyset \ \text{or}\ \ell_i' \ll \bar{\ell},\\
  \mathrm{T\mbox{-}Noi}:&\quad
  \frac{|W_{\mathrm{noise}}(t_i')|}{|W(t_i')|} \geq \alpha,\\
  \mathrm{T\mbox{-}Imb}:&\quad
  \mathrm{Var}_i(\ell_i') \ \text{or}\ \mathrm{Var}_i(\phi(t_i')) \ \text{is large}.
  \end{aligned}
\end{equation}

\textbf{Structural degradation.}
Let $A$ be the adjacency matrix, $d_i$ be node degree, and $\mathrm{sim}(i,j)$ be a semantic or label-aware similarity proxy. Structure sparsity removes valid edges, structure noise injects spurious edges, and structure imbalance creates uneven structural support:
\begin{equation}
  \begin{aligned}
  \mathrm{S\mbox{-}Spa}:&\quad E'=E\setminus E^{-}, \quad |E^{-}|/|E|\approx \alpha,\\
  \mathrm{S\mbox{-}Noi}:&\quad E'=E\cup E^{+}, \quad \mathrm{sim}(i,j)\ \text{low for }(i,j)\in E^{+},\\
  \mathrm{S\mbox{-}Imb}:&\quad
  \max_i d_i' /\min_{i:d_i'>0} d_i' \ \text{or}\ \mathrm{Var}_i(d_i') \ \text{is large}.
  \end{aligned}
\end{equation}

\textbf{Label degradation.}
Let $\mathcal{L}\subseteq V$ be the labeled training nodes, $y_i^{*}$ be the latent clean label, and $n_c$ be the number of training labels in class $c$. Label sparsity reduces supervision, label noise corrupts observed labels, and label imbalance skews class support:
\begin{equation}
  \begin{aligned}
  \mathrm{L\mbox{-}Spa}:&\quad
  \mathcal{L}'\subset\mathcal{L}, \quad |\mathcal{L}'|/|\mathcal{L}|\approx 1-\alpha,\\
  \mathrm{L\mbox{-}Noi}:&\quad
  \Pr(y_i=c\mid y_i^{*}=c')=\eta_{c'c},\\
  \mathrm{L\mbox{-}Imb}:&\quad
  \max_c n_c' /\min_{c:n_c'>0} n_c' \ \text{is large}.
  \end{aligned}
\end{equation}
These definitions are intentionally compact: they specify the target modality and the expected direction of degradation while leaving implementation details, such as exact sampling rules and compatibility constraints, to the benchmark constructor. The taxonomy therefore covers incomplete, corrupted, and uneven node text; missing, noisy, and imbalanced graph topology; and scarce, noisy, and long-tailed supervision under a shared controlled scenario space.

\begin{figure*}[!t]
  \centering
  \IfFileExists{figures/openrtag_framework_draft_v2.pdf}{%
    \includegraphics[width=\textwidth]{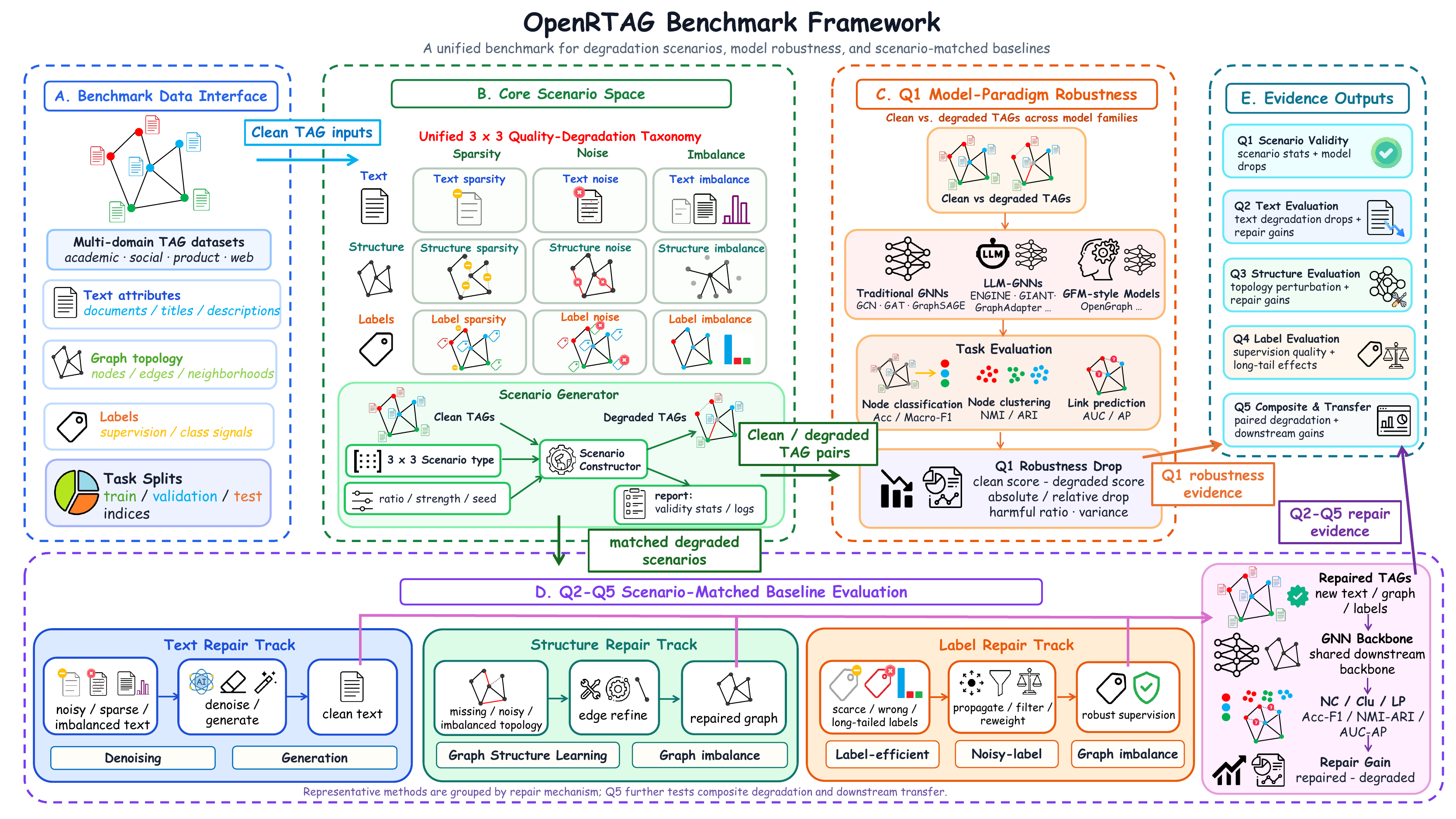}%
  }{%
    \IfFileExists{figures/openrtag_figure1_overview.png}{%
      \includegraphics[width=\textwidth]{figures/openrtag_figure1_overview.png}%
    }{%
      \fbox{%
        \parbox[c][0.24\textheight][c]{0.95\textwidth}{%
          \centering
          Figure 1 placeholder.\\
          Put the finalized overview figure at\\
          \texttt{figures/openrtag\_framework\_draft\_v2.pdf}.%
        }%
      }%
    }
  }
  \caption{Overview of the OpenRTAG benchmark framework, covering scenario construction, model-paradigm robustness, scenario-matched baselines, and evidence reporting.}
  \label{fig:openrtag-overview}
\end{figure*}

\section{OpenRTAG Benchmark Design}
OpenRTAG is organized around benchmark assets, scenario construction, evaluation coverage, and evidence reporting. Figure~\ref{fig:openrtag-overview} summarizes the resulting framework.

\subsection{Benchmark Data and Scenario Space}
OpenRTAG treats each clean TAG as a reusable source with node text, graph topology, labels, and fixed task splits. To characterize the current TAG ecosystem, we survey a broader candidate pool comprising citation and academic graphs (Cora, CiteSeer, PubMed, and Arxiv)~\cite{yang2016revisiting,hu2020ogb}, a wiki graph (WikiCS)~\cite{mernyei2020wikics}, two social graphs (Instagram and Reddit)~\cite{huang2024graphadapter,li2024glbench}, and e-commerce graphs (Children, Ratings, History, Photo, and Products)~\cite{mcauley2015image,rethinkingtag}. These datasets span small citation networks to a large academic graph with over one hundred thousand nodes and reflect recent graph-text benchmark resources~\cite{rethinkingtag,li2024glbench,feng2024taglas}. The standardized main evaluation in this paper uses nine datasets: Cora, CiteSeer, Instagram, WikiCS, PubMed, Children, Photo, History, and Arxiv. The degradation taxonomy and evaluation protocol can subsequently extend to the remaining surveyed TAG collections.

The core scenario space is the $3\times3$ taxonomy introduced in Section~2. Each scenario is instantiated by a configurable constructor that takes a clean TAG, a scenario type, and perturbation parameters such as ratio, strength, or random seed. The output is a degraded TAG with diagnostic records such as validity statistics and generation logs. Thus, scenario semantics are separated from low-level implementation details, while all generated variants remain comparable through aligned splits and task protocols.

\begin{table*}[t]
  \centering
  \scriptsize
  \setlength{\tabcolsep}{4pt}
  \renewcommand{\arraystretch}{1.12}
  \caption{Overview of model and baseline coverage in OpenRTAG. The table summarizes what is evaluated in each track; representative papers are cited in the corresponding setup paragraphs.}
  \label{tab:method-coverage}
  \arrayrulecolor{TableRule}
  \resizebox{\textwidth}{!}{%
  \begin{tabular}{>{\raggedright\arraybackslash}p{0.13\textwidth}
                  >{\raggedright\arraybackslash}p{0.19\textwidth}
                  >{\raggedright\arraybackslash}p{0.31\textwidth}
                  >{\raggedright\arraybackslash}p{0.07\textwidth}
                  >{\raggedright\arraybackslash}p{0.24\textwidth}}
    \toprule[0.16em]
    \rowcolor{TableHeader}
    \textbf{Coverage axis} & \textbf{Benchmark target} & \textbf{Representative methods} & \textbf{Track} & \textbf{Evidence produced} \\
    \midrule[0.10em]
    \rowcolor{TableBlue}
    Model paradigms & Clean-to-degraded robustness across model families & Traditional GNNs; LLM-GNN / TAG backbones; GFM-style representatives & Q1 & Scenario validity, robustness drop, harmful-case ratio \\
    \rowcolor{TableGreen}
    Text repair & Sparse, noisy, or imbalanced node text & Denoising; completion; generation; rule-based or LM-based implementation baselines & Q2 & Repair gain under text degradation \\
    \rowcolor{TableOrange}
    Structure repair & Missing, noisy, or imbalanced topology & Graph structure learning; graph purification; graph imbalance learning & Q3 & Repair gain under structural degradation \\
    \rowcolor{TablePurple}
    Label repair & Scarce, noisy, or long-tailed supervision & Label-efficient learning; noisy-label learning; imbalance-aware graph learning & Q4 & Repair gain under label degradation \\
    \rowcolor{TableBand}
    Task and composite extension & Generalization to downstream tasks and composite scenarios & Better-performing methods from Q2--Q4 & Q5 & Task-level effectiveness and performance under composite degradation \\
    \bottomrule[0.16em]
  \end{tabular}}
  \arrayrulecolor{black}
\end{table*}

\subsection{Evaluation Protocol and Coverage}
As summarized in Table~\ref{tab:method-coverage}, OpenRTAG organizes evaluation around five research questions to assess method robustness and scenario validity under controlled quality degradation. In particular, Q1 focuses on \emph{model-paradigm robustness} under controlled degradation. By comparing clean and degraded TAG pairs across traditional GNNs, LLM-GNNs, and representative GFMs, Q1 serves two purposes: it evaluates the robustness of different backbone paradigms, and it verifies whether the constructed scenarios induce meaningful, non-trivial, and consistent performance changes across model families.

Q2--Q4 evaluate \emph{scenario-matched methods} under text, structure, and label degradation, respectively. In each case, methods are matched to the degraded modality. These questions are designed to assess three aspects of method behavior: \emph{effectiveness}, namely whether a method can improve performance under its matched low-quality scenario; \emph{robustness}, namely whether its performance remains stable as the perturbation level increases; and \emph{efficiency}, including runtime, preprocessing cost, memory usage when available, and run stability. This design separates broad backbone robustness from targeted repair and mitigation performance.

Q5 further extends the evaluation beyond the main controlled setting. It examines whether the better-performing methods identified in Q2--Q4 remain effective on different downstream tasks, and how they behave when the benchmark is extended from single-scenario degradation to \emph{composite degradation settings} formed by combining two degradation scenarios. In this way, Q5 tests both the task-level generalizability of promising methods and their robustness under more realistic multi-factor degradation conditions.

\begin{figure*}[t]
  \centering
  \IfFileExists{figures/openrtag_q1_model_paradigm_heatmap_largefont_v2.pdf}{%
    \includegraphics[width=0.98\textwidth]{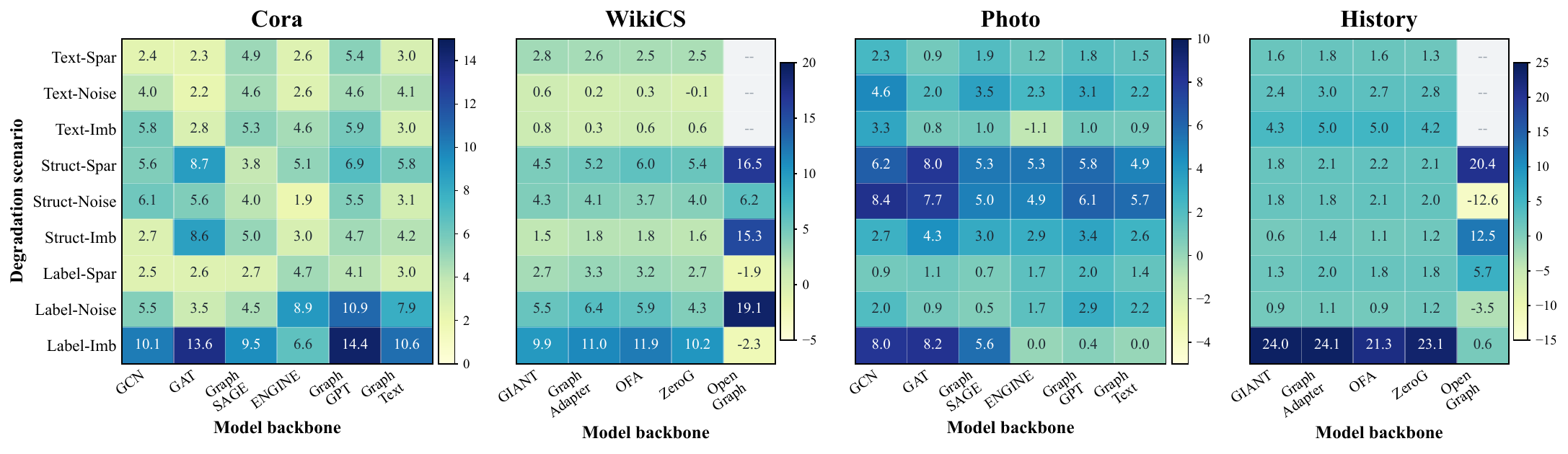}%
  }{%
    \fbox{%
      \parbox[c][0.22\textheight][c]{0.95\textwidth}{%
        \centering
        Q1 model-paradigm heatmap placeholder.\\
        Put the finalized figure at\\
        \texttt{figures/openrtag\_q1\_model\_paradigm\_heatmap.pdf}.%
      }%
    }%
  }
  \caption{Q1 model-paradigm robustness. Heatmaps report clean-to-degraded node-classification accuracy drops under the nine degradation scenarios, with color scales normalized per dataset.}
  \label{fig:q1-model-paradigm-heatmap}
\end{figure*}

\section{Experiments}
We organize the experiments around five research questions:
\textbf{Q1.} Are the degradation scenarios valid and broadly harmful across TAG backbones?
\textbf{Q2.} How effective, robust, and efficient are text-oriented baselines under text degradation?
\textbf{Q3.} How effective, robust, and efficient are structure-oriented baselines under structure degradation?
\textbf{Q4.} How effective, robust, and efficient are label-oriented baselines under label degradation?
\textbf{Q5.} Do strong baselines generalize to other downstream tasks and composite degradation scenarios?
Together, Q1 validates the scenario space, Q2--Q4 evaluate scenario-matched methods, and Q5 extends the analysis to downstream-task and composite-degradation settings.

\subsection{Q1: Scenario Validity and TAG-Backbone Robustness}
\textbf{Setup.}
Q1 aims to verify whether the proposed $3\times3$ degradation space provides valid and meaningful stress-test scenarios for TAG learning. We use node classification as the anchor task and evaluate clean and degraded TAGs over nine datasets and nine degradation scenarios. The evaluation covers three standard GNNs~\cite{kipf2017semi,velickovic2018gat,hamilton2017inductive}, seven LLM-related or broad TAG backbones~\cite{zhu2024engine,chien2022giant,tang2024graphgpt,huang2024graphadapter,zhao2023graphtext_uncertain,liu2024ofa,li2024zerog}, and a separate GFM-style compatibility analysis with OpenGraph~\cite{xia2024opengraph}.

\textbf{Results.}
The results (Fig.~\ref{fig:q1-model-paradigm-heatmap}) confirm that the constructed scenarios are valid stress tests. The internal quality statistics move in the intended directions, and the GCN anchor degrades under all nine scenarios, showing that the perturbations are both effective and non-collapsing. Label imbalance is the strongest stress case, with an average accuracy drop of 21.8 percentage points. Structure noise and sparsity are also consistently harmful, while text degradation causes milder but observable drops. Cross-backbone results further show that these degradation effects persist across traditional GNNs, LLM-GNNs, and GFM-style representatives.

\textbf{Insight.}
Overall, Q1 shows that the nine degradation scenarios are reasonable, effective, and non-trivial benchmark conditions. More importantly, the degradation effects are not limited to a single backbone: traditional GNNs, LLM-GNNs, and GFM-style representatives all exhibit different degrees of sensitivity to low-quality TAG inputs. This motivates the need for TAG robustness evaluation beyond clean-data leaderboards.

\begin{table*}[t]
  \centering
  \caption{Node-classification accuracy (\%) under text degradation with GCN as the shared backbone. Missing or failed runs are marked as OOM; the best available method for each dataset and scenario is highlighted.}
  \label{tab:text-degradation-main}
  \scriptsize
  \setlength{\tabcolsep}{3.2pt}
  \resizebox{\textwidth}{!}{%
  \begin{tabular}{llccccccccc}
    \toprule
    Scenario & Method & Cora & Citeseer & Instagram & WikiCS & PubMed & Children & Photo & History & Arxiv \\
    \midrule
    Text noise & BaseModel & 84.44{\scriptsize$\pm$0.46} & \cellcolor{gray!18}\textbf{78.06{\scriptsize$\pm$0.57}} & \cellcolor{gray!18}\textbf{65.96{\scriptsize$\pm$0.28}} & 81.70{\scriptsize$\pm$0.59} & \cellcolor{gray!18}\textbf{85.49{\scriptsize$\pm$0.36}} & \cellcolor{gray!18}\textbf{44.58{\scriptsize$\pm$0.29}} & 78.45{\scriptsize$\pm$0.11} & 79.12{\scriptsize$\pm$0.16} & \cellcolor{gray!18}\textbf{67.21{\scriptsize$\pm$0.19}} \\
     & CTD\_MLM & \cellcolor{gray!18}\textbf{84.75{\scriptsize$\pm$0.38}} & 77.59{\scriptsize$\pm$0.31} & 65.17{\scriptsize$\pm$0.58} & 81.76{\scriptsize$\pm$0.28} & 85.26{\scriptsize$\pm$0.19} & 44.24{\scriptsize$\pm$0.12} & 78.43{\scriptsize$\pm$0.06} & \cellcolor{gray!18}\textbf{79.32{\scriptsize$\pm$0.27}} & 67.16{\scriptsize$\pm$0.17} \\
     & BertDenoise & 83.70{\scriptsize$\pm$0.53} & 77.22{\scriptsize$\pm$0.71} & 65.09{\scriptsize$\pm$0.29} & 81.76{\scriptsize$\pm$0.31} & 85.44{\scriptsize$\pm$0.32} & 44.28{\scriptsize$\pm$0.18} & 78.53{\scriptsize$\pm$0.11} & OOM & OOM \\
     & RegexDenoise & 83.39{\scriptsize$\pm$0.98} & 77.64{\scriptsize$\pm$0.59} & 64.86{\scriptsize$\pm$0.62} & \cellcolor{gray!18}\textbf{82.16{\scriptsize$\pm$0.69}} & 84.95{\scriptsize$\pm$0.38} & 44.16{\scriptsize$\pm$0.90} & \cellcolor{gray!18}\textbf{78.54{\scriptsize$\pm$0.30}} & OOM & OOM \\
    \midrule
    Text sparsity & BaseModel & 85.98{\scriptsize$\pm$0.49} & 74.29{\scriptsize$\pm$0.41} & \cellcolor{gray!18}\textbf{66.46{\scriptsize$\pm$0.32}} & 79.61{\scriptsize$\pm$0.34} & 85.73{\scriptsize$\pm$0.41} & 45.65{\scriptsize$\pm$0.37} & 80.79{\scriptsize$\pm$0.20} & 80.41{\scriptsize$\pm$0.14} & 67.10{\scriptsize$\pm$0.13} \\
     & BertComplete & 85.42{\scriptsize$\pm$1.11} & 73.77{\scriptsize$\pm$0.45} & 65.40{\scriptsize$\pm$0.50} & 80.14{\scriptsize$\pm$0.69} & 85.86{\scriptsize$\pm$0.38} & \cellcolor{gray!18}\textbf{45.81{\scriptsize$\pm$0.78}} & 80.53{\scriptsize$\pm$0.03} & 80.57{\scriptsize$\pm$0.27} & OOM \\
     & NgramComplete & 85.67{\scriptsize$\pm$0.38} & 73.88{\scriptsize$\pm$0.24} & 65.30{\scriptsize$\pm$0.73} & \cellcolor{gray!18}\textbf{80.28{\scriptsize$\pm$0.56}} & 86.00{\scriptsize$\pm$0.29} & 45.71{\scriptsize$\pm$0.65} & 80.75{\scriptsize$\pm$0.26} & \cellcolor{gray!18}\textbf{80.71{\scriptsize$\pm$0.39}} & 67.03{\scriptsize$\pm$0.07} \\
     & PoDA & 84.81{\scriptsize$\pm$1.19} & \cellcolor{gray!18}\textbf{74.92{\scriptsize$\pm$0.41}} & 65.70{\scriptsize$\pm$0.50} & 79.89{\scriptsize$\pm$0.90} & 85.90{\scriptsize$\pm$0.10} & 45.77{\scriptsize$\pm$0.48} & \cellcolor{gray!18}\textbf{81.05{\scriptsize$\pm$0.30}} & OOM & OOM \\
     & UltraTAG\_S & \cellcolor{gray!18}\textbf{86.29{\scriptsize$\pm$0.56}} & 73.46{\scriptsize$\pm$0.09} & 65.80{\scriptsize$\pm$0.83} & 80.07{\scriptsize$\pm$0.73} & \cellcolor{gray!18}\textbf{86.05{\scriptsize$\pm$0.08}} & 45.78{\scriptsize$\pm$0.35} & 80.88{\scriptsize$\pm$0.21} & 80.63{\scriptsize$\pm$0.15} & \cellcolor{gray!18}\textbf{67.43{\scriptsize$\pm$0.14}} \\
    \midrule
    Text imbalance & BaseModel & 82.53{\scriptsize$\pm$1.59} & \cellcolor{gray!18}\textbf{77.85{\scriptsize$\pm$0.39}} & 75.19{\scriptsize$\pm$0.32} & 81.56{\scriptsize$\pm$0.69} & 87.32{\scriptsize$\pm$0.13} & 46.62{\scriptsize$\pm$0.39} & 79.83{\scriptsize$\pm$0.22} & 76.71{\scriptsize$\pm$0.19} & \cellcolor{gray!18}\textbf{66.74{\scriptsize$\pm$0.08}} \\
     & BertComplete & \cellcolor{gray!18}\textbf{85.49{\scriptsize$\pm$0.75}}  & 77.27{\scriptsize$\pm$0.83} & 75.65{\scriptsize$\pm$0.49} & 81.12{\scriptsize$\pm$0.70} & 87.41{\scriptsize$\pm$0.13} & 46.77{\scriptsize$\pm$0.40} & 79.75{\scriptsize$\pm$0.13} & \cellcolor{gray!18}\textbf{76.78{\scriptsize$\pm$0.31}} & OOM \\
     & NgramComplete & 84.26{\scriptsize$\pm$1.97} & 77.22{\scriptsize$\pm$0.71} & 73.93{\scriptsize$\pm$0.44} & 81.33{\scriptsize$\pm$0.27} & 87.15{\scriptsize$\pm$0.08} & 46.34{\scriptsize$\pm$0.50} & 79.67{\scriptsize$\pm$0.08} & OOM & OOM \\
     & PoDA & 84.19{\scriptsize$\pm$0.95} & 77.74{\scriptsize$\pm$0.54} & \cellcolor{gray!18}\textbf{78.56{\scriptsize$\pm$0.23}} & \cellcolor{gray!18}\textbf{82.17{\scriptsize$\pm$0.47}} & 87.52{\scriptsize$\pm$0.35} & \cellcolor{gray!18}\textbf{47.34{\scriptsize$\pm$0.07}} & \cellcolor{gray!18}\textbf{80.32{\scriptsize$\pm$0.22}} & OOM & OOM \\
     & UltraTAG\_S & 82.35{\scriptsize$\pm$0.11} & 76.02{\scriptsize$\pm$1.18} & 77.06{\scriptsize$\pm$0.18} & 80.28{\scriptsize$\pm$0.22} & \cellcolor{gray!18}\textbf{87.76{\scriptsize$\pm$0.14}} & 46.54{\scriptsize$\pm$0.43} & 79.91{\scriptsize$\pm$0.37} & OOM & OOM \\
    \bottomrule
  \end{tabular}%
  }
\end{table*}

\begin{figure}[t]
  \centering
  \includegraphics[width=\columnwidth]{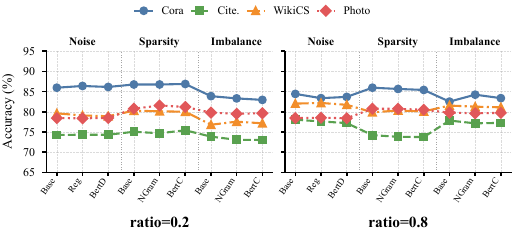}
  \caption{Q2 text-repair methods' robustness under low and high perturbation ratios.}
  \label{fig:q2-perturbation-strength}
\end{figure}

\begin{figure}[t]
  \centering
  \includegraphics[width=\columnwidth]{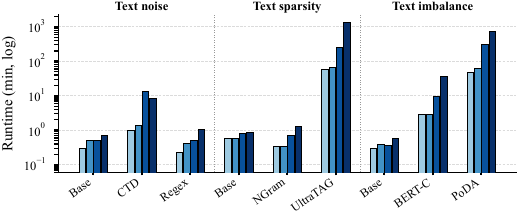}
  \caption{Runtime of representative Q2 text-oriented baselines. Bars report runtime in log-scale minutes, with different colors denoting datasets ordered as Cora, CiteSeer, WikiCS, and Photo within each method group.}
  \vspace{-10pt}
  \label{fig:q2-runtime}
\end{figure}

\begin{figure}[htbp]
  \centering
  \includegraphics[width=\columnwidth]{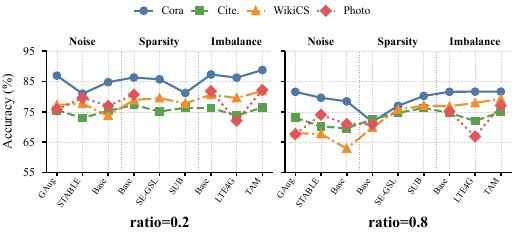}
  \caption{Q3 structure-repair methods' robustness under low and high perturbation ratios.}
  \vspace{-10pt}
  \label{fig:q3-perturbation-strength}
\end{figure}

\subsection{Q2: Effectiveness, Robustness, and Efficiency under Text Degradation}
Q2 evaluates text-oriented baselines with GCN. Three denoising methods target text noise~\cite{sun2019ctdmlm,flint2017normalization}, whereas four completion/generation methods target text sparsity and imbalance~\cite{langkilde1998ngram,wang2019poda,zhang2025ultratag}.

\textbf{Results and insights.}
Table~\ref{tab:text-degradation-main}, Figure~\ref{fig:q2-perturbation-strength}, and Figure~\ref{fig:q2-runtime} show that text-oriented repair is highly scenario- and dataset-dependent. Under text noise, denoising methods provide gains on some datasets, but the BaseModel remains competitive or best in many cases. Under text sparsity and imbalance, completion and generation better match insufficient semantic information, yet their gains remain inconsistent across datasets.

Beyond average accuracy, some methods are effective only under mild degradation and become less stable as the perturbation ratio increases. Runtime further exposes substantial preprocessing cost and OOM failures for generation- or LM-based repair. Future text-robust TAG methods should decide when and how much to intervene using degradation type, graph context, and computational budget.

\begin{figure}[t]
  \centering
  \includegraphics[width=\columnwidth]{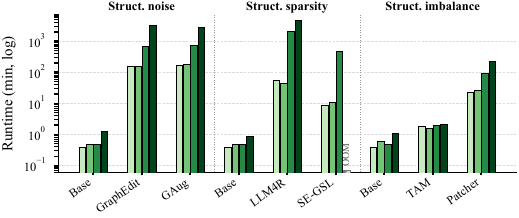}
  \caption{Runtime of representative Q3 structure-oriented baselines. Bars report runtime in log-scale minutes, with different colors denoting datasets ordered as Cora, CiteSeer, WikiCS, and Photo within each method group.}
  \label{fig:q3-runtime}
\end{figure}

\begin{table*}[t]
  \centering
  \caption{Node-classification accuracy (\%) under structural degradation with GCN as the shared backbone. For structure sparsity and structure noise, we report scenario-matched GSL subsets rather than all structure-learning methods. Missing or failed runs are marked as OOM; the best available method for each dataset and scenario is highlighted.}
  \label{tab:structure-degradation-main}
  \scriptsize
  \setlength{\tabcolsep}{3.2pt}
  \resizebox{\textwidth}{!}{%
  \begin{tabular}{llccccccccc}
    \toprule
    Scenario & Method & Cora & Citeseer & Instagram & WikiCS & PubMed & Children & Photo & History & Arxiv \\
    \midrule
    Structure noise & BaseModel & 82.29{\scriptsize$\pm$0.74} & 71.21{\scriptsize$\pm$0.45} & 65.53{\scriptsize$\pm$0.39} & 69.16{\scriptsize$\pm$0.72} & 81.82{\scriptsize$\pm$0.23} & 43.70{\scriptsize$\pm$0.43} & 74.75{\scriptsize$\pm$0.34} & 74.48{\scriptsize$\pm$0.46} & \cellcolor{gray!18}\textbf{60.43{\scriptsize$\pm$0.42}} \\
     & GAugLLM & 74.85{\scriptsize$\pm$1.57} & 68.55{\scriptsize$\pm$0.79} & 65.34{\scriptsize$\pm$0.48} & 72.97{\scriptsize$\pm$0.36} & 84.36{\scriptsize$\pm$0.17} & 43.34{\scriptsize$\pm$0.80} & 75.45{\scriptsize$\pm$0.52} & OOM & OOM \\
     & GraphEdit & \cellcolor{gray!18}\textbf{83.46{\scriptsize$\pm$0.53}} & \cellcolor{gray!18}\textbf{75.08{\scriptsize$\pm$0.72}} & 65.80{\scriptsize$\pm$0.45} & \cellcolor{gray!18}\textbf{80.01{\scriptsize$\pm$0.30}} & \cellcolor{gray!18}\textbf{87.17{\scriptsize$\pm$0.11}} & \cellcolor{gray!18}\textbf{47.92{\scriptsize$\pm$0.79}} & \cellcolor{gray!18}\textbf{78.64{\scriptsize$\pm$0.03}} & \cellcolor{gray!18}\textbf{81.17{\scriptsize$\pm$0.18}} & OOM \\
     & LLM4RGNN & 78.78{\scriptsize$\pm$0.55} & 69.07{\scriptsize$\pm$0.63} & \cellcolor{gray!18}\textbf{65.86{\scriptsize$\pm$0.54}} & 76.06{\scriptsize$\pm$0.58} & 84.81{\scriptsize$\pm$0.70} & 44.23{\scriptsize$\pm$0.24} & 71.70{\scriptsize$\pm$0.29} & 80.57{\scriptsize$\pm$0.16} & OOM \\
     & STABLE & 79.34{\scriptsize$\pm$0.80} & 72.62{\scriptsize$\pm$0.86} & 65.02{\scriptsize$\pm$0.23} & 73.37{\scriptsize$\pm$0.38} & 83.28{\scriptsize$\pm$0.22} & 46.41{\scriptsize$\pm$0.65} & OOM & OOM & OOM \\
    \midrule
    Structure sparsity & BaseModel & \cellcolor{gray!18}\textbf{82.78{\scriptsize$\pm$1.38}} & 74.40{\scriptsize$\pm$0.45} & 66.20{\scriptsize$\pm$0.37} & 74.63{\scriptsize$\pm$0.41} & 85.44{\scriptsize$\pm$0.15} & 43.53{\scriptsize$\pm$0.72} & 76.87{\scriptsize$\pm$0.11} & 79.80{\scriptsize$\pm$0.15} & \cellcolor{gray!18}\textbf{65.64{\scriptsize$\pm$0.11}} \\
     & GraphEdit & 82.04{\scriptsize$\pm$0.75} & 75.50{\scriptsize$\pm$0.39} & 65.06{\scriptsize$\pm$0.46} & \cellcolor{gray!18}\textbf{81.59{\scriptsize$\pm$0.33}} & \cellcolor{gray!18}\textbf{86.87{\scriptsize$\pm$0.12}} & \cellcolor{gray!18}\textbf{49.40{\scriptsize$\pm$0.27}} & \cellcolor{gray!18}\textbf{80.21{\scriptsize$\pm$0.38}} & OOM & OOM \\
     & LLM4RGNN & 77.80{\scriptsize$\pm$0.56} & 69.12{\scriptsize$\pm$0.57} & 63.46{\scriptsize$\pm$0.08} & 75.85{\scriptsize$\pm$0.78} & 84.01{\scriptsize$\pm$0.61} & 42.91{\scriptsize$\pm$0.59} & 73.36{\scriptsize$\pm$0.28} & \cellcolor{gray!18}\textbf{80.02{\scriptsize$\pm$0.35}} & OOM \\
     & SEGSL & 82.23{\scriptsize$\pm$0.77} & 74.19{\scriptsize$\pm$1.11} & \cellcolor{gray!18}\textbf{66.61{\scriptsize$\pm$0.44}} & 76.99{\scriptsize$\pm$1.08} & OOM & OOM & OOM & OOM & OOM \\
     & SUBLIME & 80.38{\scriptsize$\pm$1.13} & \cellcolor{gray!18}\textbf{76.12{\scriptsize$\pm$0.77}} & 65.40{\scriptsize$\pm$0.29} & 77.62{\scriptsize$\pm$0.36} & 85.39{\scriptsize$\pm$0.41} & 46.62{\scriptsize$\pm$0.12} & OOM & OOM & OOM \\
    \midrule
    Structure imbalance & BaseModel & 85.67{\scriptsize$\pm$0.38} & 75.60{\scriptsize$\pm$1.07} & \cellcolor{gray!18}\textbf{66.11{\scriptsize$\pm$0.62}} & 79.82{\scriptsize$\pm$0.28} & 87.36{\scriptsize$\pm$0.16} & 47.34{\scriptsize$\pm$0.50} & 80.37{\scriptsize$\pm$0.05} & \cellcolor{gray!18}\textbf{81.58{\scriptsize$\pm$0.20}} & 68.75{\scriptsize$\pm$0.07} \\
     & GraphPatcher & 78.91{\scriptsize$\pm$1.23} & 69.59{\scriptsize$\pm$1.65} & 59.96{\scriptsize$\pm$0.77} & 72.23{\scriptsize$\pm$1.55} & \cellcolor{gray!18}\textbf{88.87{\scriptsize$\pm$0.00}} & 48.13{\scriptsize$\pm$1.09} & 76.42{\scriptsize$\pm$0.66} & 80.13{\scriptsize$\pm$0.79} & 69.72{\scriptsize$\pm$0.31} \\
     & LTE4G & 85.42{\scriptsize$\pm$0.49} & 74.29{\scriptsize$\pm$0.31} & 60.21{\scriptsize$\pm$0.29} & 79.47{\scriptsize$\pm$0.16} & 85.44{\scriptsize$\pm$0.05} & 41.09{\scriptsize$\pm$0.33} & 71.08{\scriptsize$\pm$0.21} & 74.71{\scriptsize$\pm$0.74} & 62.22{\scriptsize$\pm$0.22} \\
     & TailGNN & 81.30{\scriptsize$\pm$7.34} & 73.72{\scriptsize$\pm$0.74} & 64.12{\scriptsize$\pm$2.20} & 79.54{\scriptsize$\pm$0.67} & 87.67{\scriptsize$\pm$1.45} & 43.96{\scriptsize$\pm$0.61} & 79.16{\scriptsize$\pm$0.65} & 80.36{\scriptsize$\pm$0.55} & 68.09{\scriptsize$\pm$0.52} \\
     & TAM & \cellcolor{gray!18}\textbf{86.78{\scriptsize$\pm$0.87}} & \cellcolor{gray!18}\textbf{76.70{\scriptsize$\pm$1.02}} & 65.80{\scriptsize$\pm$0.51} & \cellcolor{gray!18}\textbf{80.86{\scriptsize$\pm$0.68}} & 87.03{\scriptsize$\pm$0.24} & \cellcolor{gray!18}\textbf{48.39{\scriptsize$\pm$1.47}} & \cellcolor{gray!18}\textbf{81.27{\scriptsize$\pm$1.05}} & 81.54{\scriptsize$\pm$0.45} & \cellcolor{gray!18}\textbf{69.79{\scriptsize$\pm$1.49}} \\
     
    \bottomrule
  \end{tabular}%
  }
\end{table*}

\subsection{Q3: Effectiveness, Robustness, and Efficiency under Structural Degradation}
Q3 compares structure-oriented baselines under the same GCN backbone. We select scenario-matched subsets rather than applying the full GSL library to every topology failure: four relation-recovery methods for structure sparsity, four edge-purification methods for structure noise, and imbalance-oriented graph methods for structure imbalance~\cite{guo2024graphedit,llm4rgnn,zou2023segsl,liu2022sublime,li2022stable,fang2024gaugllm,ju2023graphpatcher,yun2022lte4g,liu2021tailgnn,song2022tam,chen2022topoauc}. 

\textbf{Results and insights.}
Table~\ref{tab:structure-degradation-main}, Figure~\ref{fig:q3-perturbation-strength}, and Figure~\ref{fig:q3-runtime} show that topology repair is strongly failure- and dataset-dependent. Graph editing methods are often effective under structure noise, where harmful edges need to be removed or corrected. In contrast, structure sparsity and imbalance show more mixed results: the BaseModel remains competitive on several datasets, and some repair methods even reduce performance, suggesting possible negative transfer from unreliable rewiring or reweighting. Perturbation-strength results further show that some methods lose stability under stronger degradation, while runtime results reveal substantial overhead and OOM failures for several structure-learning pipelines.

These results suggest that future structure-robust TAG methods should treat topology repair as selective intervention rather than global graph rewiring, jointly estimating which edges are trustworthy, which missing links are worth recovering, and which nodes are most vulnerable to structural defects under the downstream task.

\begin{table*}[t]
  \centering
  \caption{Node-classification accuracy (\%) under label degradation with GCN as the shared backbone. Missing or failed runs are marked as OOM; the best available method for each dataset and scenario is highlighted.}
  \label{tab:label-degradation-main}
  \scriptsize
  \setlength{\tabcolsep}{3.2pt}
  \resizebox{\textwidth}{!}{%
  \begin{tabular}{llccccccccc}
    \toprule
    Scenario & Method & Cora & Citeseer & Instagram & WikiCS & PubMed & Children & Photo & History & Arxiv \\
    \midrule
    Label noise & BaseModel & 82.90{\scriptsize$\pm$0.59} & 74.92{\scriptsize$\pm$1.10} & \cellcolor{gray!18}\textbf{64.29{\scriptsize$\pm$0.46}} & \cellcolor{gray!18}\textbf{78.91{\scriptsize$\pm$0.09}} & 85.83{\scriptsize$\pm$0.42} & 46.62{\scriptsize$\pm$0.37} & \cellcolor{gray!18}\textbf{81.07{\scriptsize$\pm$0.24}} & \cellcolor{gray!18}\textbf{81.77{\scriptsize$\pm$0.25}} & \cellcolor{gray!18}\textbf{68.56{\scriptsize$\pm$0.04}} \\
     & PIGNN & \cellcolor{gray!18}\textbf{84.62{\scriptsize$\pm$0.28}} & 74.82{\scriptsize$\pm$0.09} & 62.45{\scriptsize$\pm$0.23} & 76.76{\scriptsize$\pm$1.46} & 85.73{\scriptsize$\pm$0.29} & 46.86{\scriptsize$\pm$0.17} & 76.83{\scriptsize$\pm$1.18} & 80.99{\scriptsize$\pm$0.35} & 62.60{\scriptsize$\pm$0.76} \\
     & RNCGLN & 73.86{\scriptsize$\pm$0.21} & 73.04{\scriptsize$\pm$0.16} & 63.74{\scriptsize$\pm$0.09} & 76.11{\scriptsize$\pm$0.22} & \cellcolor{gray!18}\textbf{86.67{\scriptsize$\pm$0.19}} & \cellcolor{gray!18}\textbf{48.44{\scriptsize$\pm$0.09}} & OOM & OOM & OOM \\
     & RTGNN & 78.66{\scriptsize$\pm$1.67} & \cellcolor{gray!18}\textbf{75.03{\scriptsize$\pm$1.11}} & 64.01{\scriptsize$\pm$0.20} & 72.80{\scriptsize$\pm$1.34} & 80.73{\scriptsize$\pm$0.56} & 37.29{\scriptsize$\pm$0.76} & 68.70{\scriptsize$\pm$5.74} & 79.53{\scriptsize$\pm$0.44} & OOM \\
    \midrule
    Label sparsity & BaseModel & 85.85{\scriptsize$\pm$1.11} & \cellcolor{gray!18}\textbf{76.70{\scriptsize$\pm$0.65}} & \cellcolor{gray!18}\textbf{65.20{\scriptsize$\pm$0.41}} & 81.16{\scriptsize$\pm$0.15} & 87.65{\scriptsize$\pm$0.33} & 44.09{\scriptsize$\pm$0.55} & 82.15{\scriptsize$\pm$0.41} & 81.26{\scriptsize$\pm$0.20} & 69.35{\scriptsize$\pm$0.07} \\
     & GraFN & \cellcolor{gray!18}\textbf{88.50{\scriptsize$\pm$0.56}} & 76.18{\scriptsize$\pm$0.41} & 64.99{\scriptsize$\pm$0.34} & \cellcolor{gray!18}\textbf{84.69{\scriptsize$\pm$0.25}} & 88.70{\scriptsize$\pm$0.27} & \cellcolor{gray!18}\textbf{48.41{\scriptsize$\pm$0.40}} & \cellcolor{gray!18}\textbf{86.63{\scriptsize$\pm$0.28}} & OOM & \cellcolor{gray!18}\textbf{74.39{\scriptsize$\pm$0.44}} \\
     & GraphHop & 87.58{\scriptsize$\pm$0.11} & 75.44{\scriptsize$\pm$0.09} & 64.56{\scriptsize$\pm$0.11} & 84.17{\scriptsize$\pm$0.05} & \cellcolor{gray!18}\textbf{89.16{\scriptsize$\pm$0.03}} & 47.19{\scriptsize$\pm$0.01} & 84.46{\scriptsize$\pm$0.16} & \cellcolor{gray!18}\textbf{82.93{\scriptsize$\pm$0.04}} & OOM \\
    \midrule
    Label imbalance & BaseModel & 78.28{\scriptsize$\pm$2.59} & 74.24{\scriptsize$\pm$1.35} & 13.40{\scriptsize$\pm$2.06} & 66.59{\scriptsize$\pm$1.48} & 79.21{\scriptsize$\pm$0.40} & 25.62{\scriptsize$\pm$2.22} & 75.05{\scriptsize$\pm$0.40} & 39.69{\scriptsize$\pm$1.20} & 39.79{\scriptsize$\pm$0.27} \\
     & LTE4G & \cellcolor{gray!18}\textbf{84.46{\scriptsize$\pm$0.24}} & 76.70{\scriptsize$\pm$0.47} & 63.80{\scriptsize$\pm$0.05} & 80.23{\scriptsize$\pm$0.08} & 84.23{\scriptsize$\pm$0.11} & 39.89{\scriptsize$\pm$0.47} & 71.82{\scriptsize$\pm$0.47} & 76.05{\scriptsize$\pm$0.15} & \cellcolor{gray!18}\textbf{61.44{\scriptsize$\pm$0.19}} \\
     & TAM & 82.27{\scriptsize$\pm$1.13} & 78.82{\scriptsize$\pm$2.13} & \cellcolor{gray!18}\textbf{76.57{\scriptsize$\pm$0.37}} & \cellcolor{gray!18}\textbf{82.43{\scriptsize$\pm$0.67}} & \cellcolor{gray!18}\textbf{86.61{\scriptsize$\pm$0.06}} & \cellcolor{gray!18}\textbf{50.74{\scriptsize$\pm$2.25}} & \cellcolor{gray!18}\textbf{84.34{\scriptsize$\pm$1.00}} & \cellcolor{gray!18}\textbf{82.89{\scriptsize$\pm$0.37}} & OOM \\
     & TOPOAUC & 81.79{\scriptsize$\pm$1.84} & \cellcolor{gray!18}\textbf{79.44{\scriptsize$\pm$0.19}} & OOM & OOM & OOM & OOM & OOM & OOM & OOM \\
    \bottomrule
  \end{tabular}%
  }
\end{table*}

\begin{figure}[t]
  \centering
  \includegraphics[width=\columnwidth]{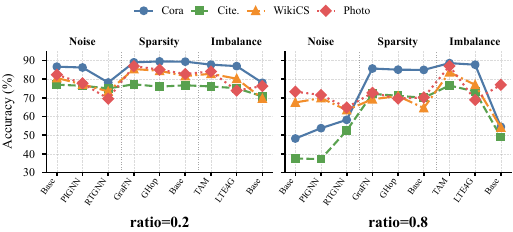}
  \caption{Q4 label-repair methods' robustness under low and high perturbation ratios.}
  \label{fig:q4-perturbation-strength}
\end{figure}

\begin{figure}[t]
  \centering
  \includegraphics[width=\columnwidth]{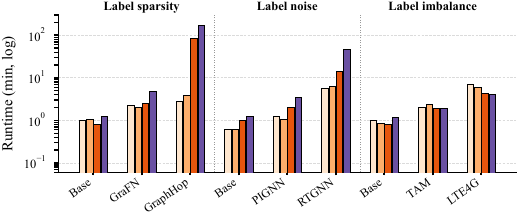}
  \caption{Runtime of representative Q4 label-oriented baselines. Bars report runtime in log-scale minutes, with different colors denoting datasets ordered as Cora, CiteSeer, WikiCS, and Photo within each method group.}
  \label{fig:q4-runtime}
\end{figure}

\begin{figure}[!htbp]
  \centering
  \includegraphics[width=\columnwidth]{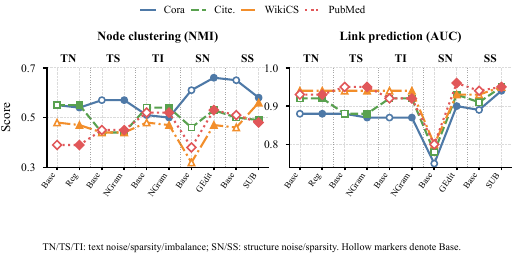}
  \caption{Q5 Performance under different downstream tasks.}
  \label{fig:q5-downstream-task}
\end{figure}

\subsection{Q4: Effectiveness, Robustness, and Efficiency under Label Degradation}
Q4 evaluates label-oriented baselines under scarce, corrupted, and long-tailed supervision. We use two label-sparsity methods, three noisy-label methods, and four imbalance-oriented methods~\cite{lee2022grafn,xie2021graphhop,du2023pignn,zhu2024rncgln,qian2023rtgnn,yun2022lte4g,liu2021tailgnn,song2022tam,chen2022topoauc}. 

\textbf{Results and insights.}
Table~\ref{tab:label-degradation-main}, Figure~\ref{fig:q4-perturbation-strength}, and Figure~\ref{fig:q4-runtime} show that supervision failures differ substantially across failure modes. Label imbalance causes severe drops on several datasets, and imbalance-aware methods such as TAM and LTE4G can bring large gains when long-tailed supervision dominates. Label sparsity also benefits from label-efficient methods on many datasets, although the best method varies between GraFN and GraphHop. In contrast, label noise is more difficult to repair: the BaseModel remains competitive in many cases, suggesting that explicit label correction may yield unstable gains when corrupted labels are hard to identify. The perturbation-strength results further show that supervision-oriented methods are sensitive to degradation severity, while runtime results indicate that they are generally more lightweight than text or structure repair methods, despite some costly or failed runs.

These results suggest that future supervision-robust TAG methods should move beyond fixed label correction, propagation, or reweighting, and instead learn uncertainty-aware supervision policies that decide when to trust, correct, propagate, or rebalance labels under different forms of supervision failure.

\begin{figure}[!htbp]
  \centering
  \includegraphics[width=\columnwidth]{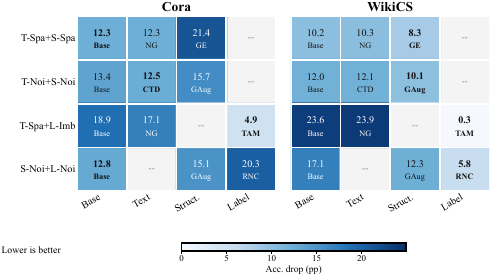}
  \caption{Q5 Performance under composite scenarios.}
  \label{fig:q5-composite-scenarios}
\end{figure}

\subsection{Q5: Extension to Composite Scenarios and Downstream Tasks}

Q5 extends the main node-classification analysis to harder and broader settings. We first construct four two-factor composite scenarios by combining text, structure, and label degradation, and compare the GCN with the best matched baselines selected from Q2--Q4. We further evaluate representative text and structure scenarios on node clustering and link prediction to examine whether the observed robustness patterns generalize beyond node classification.

\textbf{Results and insights.}
Figures~\ref{fig:q5-downstream-task} and~\ref{fig:q5-composite-scenarios} show that the proposed degradation settings remain informative beyond node classification. Similar performance gaps appear in node clustering and link prediction, indicating that OpenRTAG captures quality issues that affect TAG learning across tasks. Under composite degradation, existing methods show less stable gains than in single-scenario settings, suggesting that repairing one modality alone may be insufficient when text, structure, and label quality issues coexist.
These results suggest that future robust TAG methods should move beyond one-defect-at-a-time repair and jointly optimize text, topology, and supervision signals under multi-factor degradation.

\section{Conclusion}
We present \textbf{OpenRTAG}, a robustness benchmark for text-attributed graph learning under low-quality data. OpenRTAG defines a unified $3\times3$ degradation space across text, structure, and labels, covering sparsity, noise, and imbalance, and evaluates representative models and baselines across multiple datasets and tasks. The results show that TAG robustness is scenario-dependent and that existing methods remain limited under strong or composite degradation, motivating future methods that jointly address text, topology, and supervision quality. A limitation is that OpenRTAG focuses on representative degradation settings, leaving more dynamic and domain-specific quality issues for further investigation in future work.

\bibliography{openrtag_refs}

\end{document}